\documentclass{article}

\usepackage[preprint]{neurips_2023}

\usepackage[utf8]{inputenc} 
\usepackage[T1]{fontenc}    
\usepackage[hidelinks]{hyperref}       
\usepackage{url}            
\usepackage{booktabs}       
\usepackage{amsfonts}       
\usepackage{nicefrac}       
\usepackage{microtype}      
\usepackage{xcolor}         
\usepackage{graphicx}
\usepackage{amsmath}
\usepackage{mathrsfs}

\title{Predictive variational autoencoder for learning robust representations of time-series data}

\author{%
  Julia H. Wang \\
  Cold Spring Harbor Laboratory School of Biological Sciences\\
  Cold Spring Harbor Laboratory\\
  Cold Spring Harbor, New York, USA \\
  \texttt{julwang@cshl.edu} \\
  \And
  Dexter Tsin \\
  Princeton Neuroscience Institute \\
  Prineton University \\
  Princeton, New Jersey, USA \\
  \texttt{dexter.tsin@princeton.edu} \\
  \AND
  Tatiana A. Engel \\
  Princeton Neuroscience Institute \\
  Prineton University \\
  Princeton, New Jersey, USA \\
  \texttt{tatiana.engel@princeton.edu} \\
}

\begin{document}

\maketitle

\begin{abstract}
Variational autoencoders (VAEs) have been used extensively to discover low-dimensional latent factors governing neural activity and animal behavior. However, without careful model selection, the uncovered latent factors may reflect noise in the data rather than true underlying features, rendering such representations unsuitable for scientific interpretation. Existing solutions to this problem involve introducing additional measured variables or data augmentations specific to a particular data type. We propose a VAE architecture that predicts the next point in time and show that it mitigates the learning of spurious features. In addition, we introduce a model selection metric based on smoothness over time in the latent space. We show that together these two constraints on VAEs to be smooth over time produce robust latent representations and faithfully recover latent factors on synthetic datasets. 
\end{abstract}

\section{Introduction}
One central goal of neuroscience is to understand the link between neural activity in the brain and behavior. 
Both neural activity and behavior are often measured via high-dimensional observables but are governed by low-dimensional latent factors. Identifying these factors from data can reveal the correspondence between the neural activity and behavior leading to new scientific hypotheses (\cite{cunningham_dimensionality_2014}).

This latent to observable mapping can be modeled as a generative process in variational autoencoders (VAE), and many have leveraged VAEs to uncover latent factors from animal behavior and neural recordings (\cite{pandarinath_inferring_2018, zhou_learning_2020, pei_neural_2021, petersonUnsupervisedDiscoveryFamily2023, luxem_identifying_2022}). However, to reliably interpret latent representations, we need to ensure that they capture some true underlying structure of the data and do not contain spurious features resulting from noise. For example, if different instances of a model produced different representations from the same data, it would be impossible to meaningfully interpret these representations. Thus, robustness and reproducibility are prerequisites for interpretability.

However, standard VAEs do not produce robust representations (\cite{locatello_challenging_2019, keshtkaran_enabling_2019, radhakrishnan_memorization_2019}), indicating that these models can learn spurious features unrelated to underlying structure of the data. There are two possible causes of how VAEs may learn spurious features. First, the VAE might learn a functional mapping that is irrelevant to the low-dimensional structure but sufficient for reconstructing the data (\cite{keshtkaran_enabling_2019, radhakrishnan_memorization_2019}). The reconstruction goal of VAEs only requires the model to separate dissimilar points but does not incentivize the model to keep similar points close in the latent space. Thus, there may be several local optima in the loss landscape at which the VAE can map points in the latent space to points in the original space without learning the correct low-dimensional structure.

Second, VAE can overfit to noise in the data by trying to separate data points based on the noise features. Traditional practice to avoid overfitting is regularization, which involves selection of hyperparameters by optimizing the model's performance on a held-out validation dataset. However, flexible machine learning models with many parameters can generalize well on unseen data despite learning to interpolate perfectly through noise in the training data (\cite{belkin_reconciling_2019, bartlett_benign_2020}). In these instances, many models can generalize equally well but have different learned features and therefore interpretations (\cite{genkin_moving_2020}). This type of overfitting is not an issue if generalization performance is the only goal, but it undermines the interpretation of the latent representations learned by the model.

Solutions proposed to avoid the learning of spurious features involve 1) adding inductive bias to the model architecture or through regularization (\cite{locatello_challenging_2019, khemakhem_variational_2020}) and 2) using appropriate model selection metrics to choose robust features that are reproducible across different model instances (\cite{duan_rl2_2016, genkin_moving_2020}). Here we extend these ideas to learning robust low-dimensional representations of time series data with VAEs. We first show that standard VAEs are prone to learning spurious features. We then introduce an inductive bias in the VAE architecture and a model selection metric which both promote smoothness of latent factors over time. The temporal smoothness of latent factors naturally arises from the continuous dynamics inherent in the vast majority of physical and biological processes. Therefore, it serves as a natural prior for time-series data generated by these processes. We show that with these changes, VAEs learn robust representations on synthetic and biological datasets.

\section{Related Work}

\subsection{Learning interpretable representations}
Uncovering robust low-dimensional representations of biological data with deep learning models allows for downstream analysis and interpretation to derive new scientific hypotheses. The goal of achieving robust representations directly aligns with efforts in developing identifiable models. A model is \emph{identifiable} if its parameterization of a certain set of observations is unique (\cite{khemakhem_variational_2020}). The standard VAE is not identifiable itself, but some modifications of the VAE do achieve identifiability, for example, through conditioning on an additionally observed auxillary variable (\cite{khemakhem_variational_2020, zhou_learning_2020, mita_identifiable_2021}). 

Learning robust or identifiable representations is distinct from achieving disentanglement in VAEs (\cite{khemakhem_variational_2020}). Disentanglement seeks to create a representation that separates meaningful features of the high-dimensional data into independent generative factors in the latent space. However, the goals of identifiability in representation learning and disentanglement are related, as many of the modifications to VAE architecture and learning paradigms that produce robustness or identifiability are also useful in achieving disentanglement (\cite{mita_identifiable_2021}).

Various modifications to VAEs have been explored, both in and outside of neuroscience, to achieve more robust and interpetable representations. Some of these variations promote smoothness over time or spatial dimensions (\cite{gulrajani_pixelvae_2016, klindt_towards_2021}). Weak supervision and self-supervision are also often explored in pursuit of robust representation learning and identifiability. Weak supervision often leverages multiple data modalities for regularization by introducing an auxillary variable, such as activity of another neuron or behavioral measurements (\cite{mita_identifiable_2021, zhou_learning_2020, liu_drop_2021, azabou_mine_2021}). However, not all datasets have associated auxillary variables, so such a method is unsuitable in those cases. Self-supervised learning involves training the model to predict an augmented version of the original data, called a self-supervised label. Doing so ensures that the model learns to preserve features that are shared between the original and augmented data. For example, data augmentation can involve using a semantic-preserving transformation of an image (such as a rotation) (\cite{sinha_consistency_2021}) or dropping and swapping spikes in a neural activity dataset (\cite{liu_drop_2021}). These kinds of data augmentation are specific to the dataset at hand, and it is generally not obvious which transformations preserve the true latent structure for every data type. 

Some non-variational model architectures also seek to learn robust representations of time-series data, such as contrastive predictive coding (\cite{oord_representation_2019}) or prediction of neighboring words (\cite{mikolov_efficient_2013}). However, representation learning with predictive coding models has not been studied in terms of identifiability and robustness of representation, nor have these models been systematically compared to VAEs in this area.

\subsection{Model selection metrics for interpretability}

Even with inductive biases added to promote robustness or identifiability, there is still a need for model selection metrics to avoid overfitting to noise in the training and validation data. A standard metric for model selection in unsupervised settings is cross-validated performance, such as reconstruction loss for VAEs. However, since overparametrized models can achieve high validation performance despite learning spurious features (\cite{belkin_reconciling_2019, genkin_moving_2020}), there is a need for validation metrics that directly assess the robustness of features learned by the model. One approach is to evaluate the model performance in predicting an external variable (\cite{higgins_-vae_2017, pei_neural_2021}); however, this relies on some external label rather than evaluating the quality of latent representations directly. Another approach measures the similarity of latent representations learned by different model instances, as true features should be similar, whereas spurious features are not reproducible (\cite{duan_unsupervised_2020,genkin_moving_2020}). 

\section{Methods}

\subsection{Datasets}

To study how standard VAEs learn spurious features and to test how our approach overcomes this problem, we used synthetic data with known low-dimensional structure, as well as biological neural recording data. Synthetic datasets are useful because we know the ground-truth latent structure used to generate the data, although they may not capture the entire complexity of real biological data. 

\textbf{Hidden Markov Model (HMM) and Gaussian clusters}: We generated synthetic data points from a three-state HMM with emissions sampled from 3 clusters mimicking wake, rapid eye movement (REM), and slow wave sleep (SWS) brain states. Each cluster was modeled as a Gaussian distribution in 31-dimensional space, where dimensions represent local field potential (LFP) power in 30 frequency bands and total power of the electromyography (EMG) signal with mean and variance of the Gaussian distributions matched to the biological data. The transition dynamics were also matched to wake, REM and SWS states in biological data. In Fig.~\ref{fig2}, we use this dataset without considering time information, essentially treating data-points as independent samples from the 3 high-dimensional Gaussian clusters.

\textbf{Spiral}: We sampled data points uniformly along a two-dimensional spiral, with time index increasing monotonically along the spiral starting from the innermost point. We then nonlinearly embedded these points into 31-dimensional space and added Gaussian white noise.

\textbf{Visual cortex LFP data}:
We used combined EMG and LFP data recorded from the visual cortex of a mouse during continuous 24-hour recordings over 12 days (\cite{soltani_sleepwake_2019}). Each data point represents a 2 second time window. For this 2-second time window, we extracted the LFP power in 30 frequency bands and total EMG power to form a 31-dimensional feature vector representing the signals. A subset of data points had expert-provided labels of wake, REM and SWS brain states.

\subsection{Time-Neighbor VAE architecture}

\begin{figure}[ht]
\begin{center}
\centerline{\includegraphics{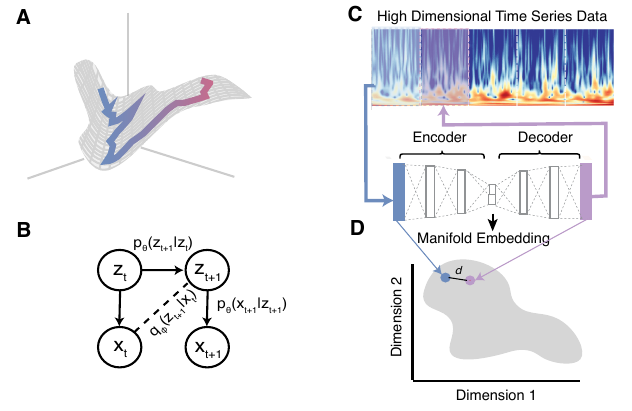}}
\caption{
\textbf{A}, Generative model. The data arises from latent dynamics (illustrated as a blue-to-pink trajectory) evolving on a low-dimensional manifold (2-D grey surface). This manifold is nonlinearly embedded in a high-dimensional space of observations, in which each axis represent one feature in the data.
\textbf{B}, Directed graphical model. Solid lines indicate the generative model and dashed lines indicate the variational approximation.
\textbf{C}, Neighbor VAE Architecture. High-dimensional time series data is binned to desired time windows and represented as a high-dimensional vector. This vector is passed into a VAE that predicts the vector for the next point in time. 
\textbf{D}, The distance between neighboring points in time on the manifold embedding discovered by the VAE is used as a metric for model selection. 
} 
\label{fig1}
\end{center}
\vskip -0.2in
\end{figure}

We added an inductive bias to the VAE architecture that promotes learning of underlying low-dimensional structure in time-series data. We assume that latent factors evolve smoothly in time as a Markov process over the latent space (Fig.~\ref{fig1}A,B). Thus, we constrained the VAE architecture to be autoregressive by modifying the objective from reconstructing the original data point to predicting the next point in time (Fig.~\ref{fig1}B,C). In this way, the function the model parameters learn encompasses both the generative function that maps low-dimensional latent factors to observables and the transition probability over time between points in the latent space. Thus, we minimize the following loss function (derivation in Appendix: A.1):
\begin{equation}
\mathscr{L} =  -\textrm{E}_{z \sim q_\phi(z_{t+1}|x_t)}[\log p_\theta(x_{t+1}|z_{t+1})] + \beta \textrm{KL}(q(z_{t+1}|x_t)\|p(z_{t+1})) .
\end{equation}
Here $x_{t}$ is the data point at time $t$ and $z_{t}$ is its latent representation. $p_\theta(x_{t+1}|z_{t+1})$ is the likelihood of $x_{t+1}$ conditioned on $z_{t+1}$ under generative model $p_\theta$, and $q_\phi(z_{t+1}|x_t)$ is the approximated posterior of the future latent state $z_{t+1}$ given the observation $x_t$. 
We allowed for a weight on the KL divergence term $\beta$, which was proposed as the $\beta$-VAE (\cite{higgins_-vae_2017}).
We call our autoregressive VAE the Time-Neighbor VAE (TN-VAE).

\subsection{Model selection metric: Neighbor Loss}
To prevent learning spurious features due to overfitting to noise in the training and validation data,
we defined a model selection metric based on our assumption that latent factors evolve smoothly over time in the latent space. For each model instance, we calculated the distance in the latent space between the representations of each data point and the next point in time, normalized by the overall size of the latent manifold (Fig.~\ref{fig1}D). This metric which we termed Neighbor Loss (NL) was defined as:
\begin{equation}
\textrm{NL} = \sum_{t=0}^{N-1}\frac{\|z_{t+1} - z_{t}\|}{\bar{z}},
\end{equation}
where $\bar{z} = \sum_{t=0}^{N} \| z_t \| / N$ is the average distance from the origin across latent representations of all $N$ points in the dataset. Theoretically, if transitions between points in the latent space follow a Gaussian random walk, then minimizing the absolute distance between latent representations of neighboring points in time is equivalent to maximizing their log-likelihood (Appendix: A.2). Calculating NL loss has a small computational cost scaling linearly with the number of data points, thus calculating NL throughout training adds only a negligible computational overhead.

\subsection{Training details}

We trained large ensembles of VAE and TN-VAE models across a comprehensive range of hyperparameters. We compiled a range of values across $5$ hyperparameters: number of layers in the model ($2 - 4$), dimensions of each layer ($50 - 400$), KL-divergence weight ($10^{-4} - 10^{-3}$), batch size (spiral dataset: $128 - 1024$; HMM and LFP datasets: $2048 - 8192$), and learning rate ($10^{-5} - 10^{-3}$). We used multiple unique seeds (spiral dataset: $5$; HMM and LFP datasets: $3$) that determined the partition of data into training and validation splits and model parameters at initialization. We fitted all models for $500$ epochs, at which point the validation loss curves fully plateaued. In total, we trained multiple model instances per dataset (spiral dataset: $1440$; HMM and LFP datasets: $648$) for subsequent analysis.

\subsection{Metrics for the quality of learned representations}
We used silhouette score with respect to the ground-truth clusters to quantify how well the latent representations recover the true low-dimensional structure in the data. The silhouette score for a single sample is defined as:
\begin{equation}
s(i) = \frac{b(i) - a(i)}{\textrm{max}{[ a(i), b(i) ] }},
\end{equation}
where $a(i)$ is the mean distance between point $i$ and all other points in the same cluster, and $b(i)$ is the mean distance between point $i$ and all points from the other nearest cluster. The silhouette score of the whole dataset is the mean over silhouette score of all samples. We use the ground-truth clusters for the HMM dataset, and the expert labels of wake, SWS, and REM states for the LFP dataset. For the spiral dataset, we define ground-truth clusters as groups of 100 consecutive data points along the spiral.

We quantified the robustness of representations learned by different model instances using procrustes distance between their encodings of the same test data. Generalized procrustes distance is the minimum Euclidean distance between two data matrices $\mathbf{A} \in \mathbb{R}^{n \times d}$ and $\mathbf{B} \in \mathbb{R}^{n \times d}$ ($n$: number of data points, $d$: number of dimensions) subject to any rotation, translation, scaling and reflection (\cite{gower_generalized_1975}).

\section{Results}

\subsection{VAE learns spurious features}
\begin{figure}[ht]
\begin{center}
\centerline{\includegraphics[width=\columnwidth]{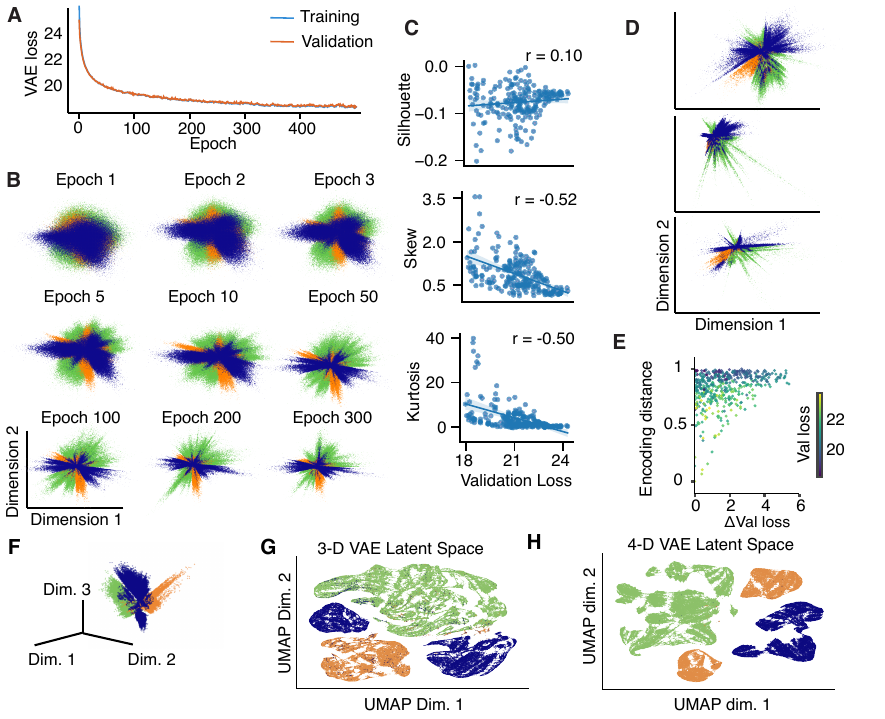}}
\caption{
\textbf{A}, Training and validation loss for an example VAE trained on the Gaussian clusters dataset.
\textbf{B}, Latent representations learned by this VAE at different training epochs.
\textbf{C}, Validation loss versus silhouette score (upper panel), skew (middle panel) and kurtosis (lower panel) of latent encodings of the ground-truth clusters across 288 VAEs with unique hyperparameters combinations. 
\textbf{D}, Latent representations after 500 epochs in 3 VAEs with different hyperparameters with similar validation loss show lack of robustness.
\textbf{E}, Models with low, more similar validation loss did not necessarily have more similar latent encodings as measured by procrustes distance (each dot is a model pair, a random sample of 500 model pairs is shown). Color indicates average validation loss between two models.
\textbf{F}, Latent representations in a three-dimensional VAE with the best validation loss across 288 models with different hyperparameters trained on the same data.
\textbf{G}, UMAP-projection of the three-dimensional latent space from F. 
\textbf{H}, Same as G for the VAE with four latent dimensions that had the best validation loss.}
\label{fig2}
\end{center}
\vskip -0.2in
\end{figure}

To illustrate that standard VAEs are prone to learning spurious features, we trained VAEs on a synthetic dataset of points drawn from 3 high-dimensional Gaussian clusters. Both the training and validation loss decreased monotonically throughout 500 training epochs (Fig.~\ref{fig2}A). Thus, the standard approach would be to select the model at the last training epoch that has the lowest validation loss. However, despite good reconstruction performance on validation data, the VAE not only failed to correctly separate the 3 clusters in the learned latent representation, but also learned spurious features. Over training epochs, the latent representation gradually lost the smooth Gaussian shape and feathered into streak-like patterns. (Fig.~\ref{fig2}B). To quantify how well a model separates the three clusters, we computed the silhouette score of the latent encodings using the ground-truth cluster labels. Across 288 VAEs trained with different hyperparameter combinations, there was not a single model that separated clusters well, and the silhouette score did not correlate with the validation loss (Fig.~\ref{fig2}C). In addition, models with lower validation loss had higher per-cluster skew and kurtosis of the latent encodings, indicating deviation from Gaussian distribution (Fig.~\ref{fig2}C). These findings show that lower validation loss does not imply a more accurate representation of the ground-truth structure in the latent encodings.

Furthermore, separate model instances trained on different splits of the data achieved similar validation loss but uncover different latent representations (Fig.~\ref{fig2}D), indicating that the VAE learns spurious features that are not robust. We quantified similarity of latent representations between a pair of models using the encoding distance, defined as procrustes distance (\cite{gower_generalized_1975}) between their latent encodings of the same test data. The encoding distance measures the disparity between two latent representations of the data allowing for arbitrary scaling, dilation, rotation, and reflection of the latent space. One might expect that models with lower and more similar validation loss produce more similar encoding spaces. However, we observed the opposite behavior: models with more similar, low validation loss have more dissimilar encoding spaces as indicated by higher encoding distance (Fig.~\ref{fig2}E).

One concern may be that VAEs learn spurious features because two latent dimensions are not sufficient to faithfully represent the ground-truth structure in the data. Therefore, it is common to fit VAEs with a high-dimensional latent space and then project the latent encodings to two dimensions for visualization using Uniform Manifold Approximation and Projection (UMAP) (\cite{mcinnes_umap_2020}) or t-distributed Stochastic Neighbor Embedding (tSNE) (\cite{maaten_visualizing_2008}). However, we found that VAEs with three latent dimensions learn similar spurious features as two-dimensional VAEs  (Fig.~\ref{fig2}F). Moreover, UMAP-projection of higher-dimensional latent spaces produced spurious structures with more clusters than the ground truth (Fig.~\ref{fig2}G,H), consistent with the fact that UMAP and tSNE methods are not guaranteed to preserve local or global distances, are sensitive to the choice of hyperparameters, and thus can produce spurious features (\cite{chari_specious_2023}). These results demonstrate that variational autoencoders are not suitable for faithfully identifying latent structure in high-dimensional data, and validation loss is not a suitable metric for selecting interpretable models with robust features.

\begin{figure}[ht]
\begin{center}
\centerline{\includegraphics[width=\columnwidth]{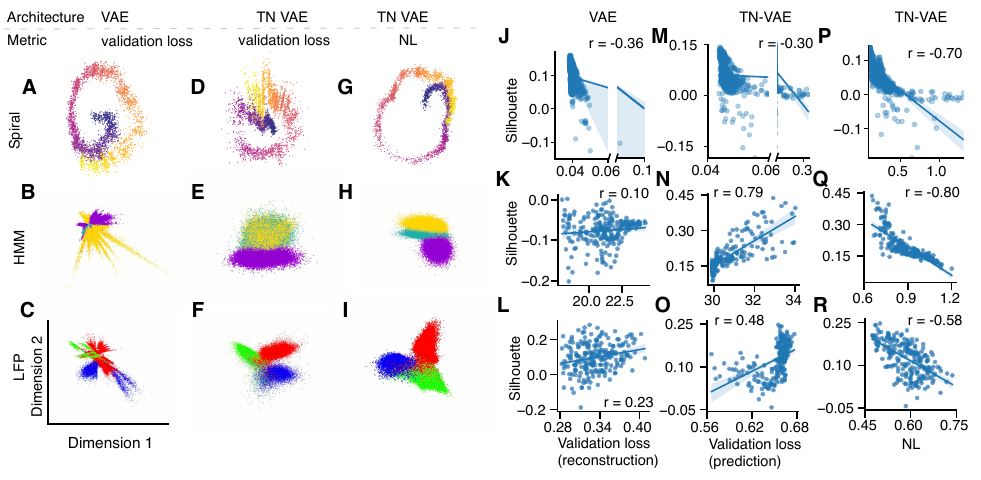}}
\caption{Latent representations on a held-out test set learned by the standard VAE with validation-loss model selection (\textbf{A-C}), TN-VAE with validation-loss model selection (\textbf{D-F}), and TN-VAE with NL model selection (\textbf{G-I}) for three datasets: synthetic spiral (\textbf{A, D, G}, points colored by the position on the original spiral), synthetic HMM (\textbf{B, E, H}, points colored by the 3 HMM states), and biological LFP data (\textbf{C, F, I}, points colored by expert-provided labels of wake, REM, and SWS brain states). We quantify the correspondence between latent encodings and the ground truth using the silhouette score of encodings with respect to clusters formed by each 100 consecutive data points for the spiral dataset (\textbf{J, M, P}), silhouette score of encodings with respect to the ground-truth clusters for the HMM dataset (\textbf{K, N, Q}), and silhouette score of encodings with respect to the expert-labeled wake, REM, and SWS states for the LFP dataset (\textbf{L, O, R}). For the VAE, lower validation loss does not correlate with the higher silhouette score (\textbf{J, K, L}). For the TN-VAE, validation loss is only slightly negatively correlated with silhouette score for spiral (\textbf{M}) and positively correlated with silhouette score for the HMM and LFP datasets (\textbf{N, O}). In contrast, lower NL corresponded to higher silhouette score for all datasets (\textbf{P, Q, R}), indicating better correspondence of latent encodings with the ground truth. 
} 
\label{fig3}
\end{center}
\vskip -0.2in
\end{figure}

\subsection{Priors on smoothness over time promote learning of true and not spurious features}

We introduce a modified model architecture TN-VAE and a NL model selection metric which promote smoothness of latent factors over time (Methods). We used two synthetic datasets with known low-dimensional structure (Fig.~\ref{fig3}A-H) to test how these modifications affect the quality of learned latent representations in comparison to the standard VAE and standard model selection based on validation loss. First, the standard VAE with the validation-loss model selection did not necessarily recover the correct low-dimensional structure without learning spurious features (Fig.~\ref{fig3}A,B, Figs.~\ref{figs1}A,~\ref{figs2}A). The level and pattern of overfitting (to noise or by memorization) depends on the noise level in the data observations (Fig.~\ref{figs4}). TN-VAE introduces an inductive bias that pushes the model to learn meaningful features, but can still overfit to noise in the data or fail to select the best model (Fig.~\ref{fig3}D,E, Figs.~\ref{figs1}B,~\ref{figs2}B). Using TN-VAE in combination with NL model selection metric resulted in the most faithful reconstruction of the original low-dimensional structure in the data without learning spurious features (Fig.~\ref{fig3}G,H, Figs.~\ref{figs1}C,~\ref{figs2}C). 

We then applied each approach to biological LFP data recorded in the visual cortex of a mouse over normal day and night activity. While the ground-truth structure in these data is unknown, the domain knowledge suggests the existence of 3 major clusters in these data corresponding to wake, REM, and SWS brain states. The standard VAE with best validation loss failed to separate the clusters at all and instead learns spurious features (Fig.~\ref{fig3}C, Fig.~\ref{figs3}A). The TN-VAE with validation-loss model selection separates the clusters (Fig.~\ref{fig3}F) but does not reliably achieve consistency between latent encodings of the models with the best validation loss (Fig.~\ref{figs3}B). Finally, TN-VAE with the NL model selection metric produced a smooth latent representation with 3 clusters that matched the human-expert labeling of wake, REM, and SWS brain states (Fig.~\ref{fig3}I), which was highly consistent across the models with the best NL (Fig.~\ref{figs3}C).

Across all models trained on these three datasets, lower validation loss did not necessarily indicate better correspondence between latent encodings and the ground truth. We quantified this relationship using silhouette score of encodings with respect to clusters formed by each 100 consecutive data points for the spiral dataset (Fig.~\ref{fig3}J,M), the silhouette score of encodings with respect to the ground-truth clusters for the HMM dataset (Fig.~\ref{fig3}K,N), and the silhouette score of encodings with respect to expert-labeled wake, REM, and SWS states in the LFP dataset (Fig.~\ref{fig3}L,O). In contrast, lower NL in the TN-VAE model was systematically correlated with better correspondence between the latent encodings and the ground truth, indicated by higher silhouette score (Fig.~\ref{fig3}P,Q,R). Thus, the TN-VAE model architecture and NL model selection metric together are most effective in recovering latent encodings representing ground truth structure.

\subsection{Neighbor Loss metric selects models with robust representations}

\begin{figure}[ht]
\begin{center}
\centerline{\includegraphics[width=\columnwidth]{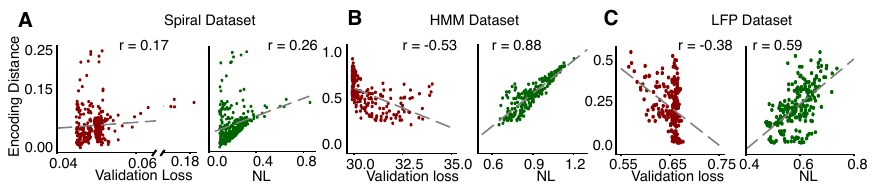}}
\caption{For each hyperparameter combination, we trained multiple model instances (spiral dataset: $1440$, HMM and LFP datasets: $648$) on different training and validation splits. We measured the encoding distances between latent representations learned by each pair of model instances on a test set. We also calculated the average validation loss and average validation NL over the model instances for each hyperparameter combination. NL correlated more with encoding distance than validation loss for synthetic spiral (\textbf{A}) HMM (\textbf{B}) and LFP (\textbf{C}) datasets.}
\label{fig4}
\end{center}
\vskip -0.2in
\end{figure}

Finally, we tested how the Neighbor Loss (NL) model selection metric corresponds to the robustness of the learned representations. For all three datasets, we chose a variety of different hyperparameter configurations, and for each configuration, we trained multiple model instances with unique training and validation data splits. Each model instance produces a latent representation of the same held-out test dataset. We then measured the average encoding distance between latent representations of the test set for each pair of model instances across a set of unique seeds. Lower encoding distance indicates more consistency in representations across model instances and thus, can be used to assess the robustness of learned latent representations. We found that NL correlates well with encoding distance, whereas validation loss was not correlated with how consistent the latent representations are across model instances (Fig.~\ref{fig4}). Intuitively, if models underfit, there would be no correlation between the structure of the latent representation and time information. Vice versa, overfitted models separate neighboring points in time based on noise features that are random across training instances. As a result, minimizing the NL metric as a criterion for model selection leads to the selection of robust latent representations.

\section{Discussion}
Achieving similar representations across different models trained on the same biological data is a prerequisite to scientific interpretation of such representations. We tested how and when this similarity can fail and degenerate representations can arise. Specifically, we seek to understand overfitting to noise in VAE latent representations of high-dimensional neural data and to uncover strategies to mitigate such overfitting. We show that the representations produced by the standard VAE with validation-loss model selection often lack robustness and thus cannot be reliably interpreted. We leverage smoothness of latent representations over time to introduce both an inductive bias on model architecture and a model selection metric, which together prevent learning of spurious features and lead to robust representations. 

The introduction of time as a regularization in neural networks has been used previously, both in variational autoencoder architectures (\cite{schneider_learnable_2023, liu_drop_2021, sedler_expressive_2023}) and more broadly in feedforward networks (\cite{toosi_brain-like_2023}) and transformer architectures (\cite{devlin_bert_2019}). Here, we extend the use of time to mitigate overfitting and learning spurious features in VAEs. In addition, we propose its use as a model selection metric, which unsupervised methods such as VAEs often lack. In contrast to current model selection metrics for VAEs (\cite{higgins_-vae_2017, pei_neural_2021, duan_unsupervised_2020}), our proposed NL metric is intrinsic to the latent representations themselves, not requiring external labels, and also can be used during training unlike methods that need to compare large ensembles of trained models. Thus, our contributions are to 1) show that standard VAEs are prone to learning spurious features; 2) show how using time in VAE architecture is an effective inductive bias and regularizer and 3) introduce a novel model selection metric based on smoothness over time of latent representations.

One way to recover meaningful and interpretable representations of data by VAEs is through achieving identifiability. It is important to note that true identifiability cannot be achieved without an external label (\cite{locatello_challenging_2019}), and our methods does not necessarily achieve identifiability. Instead, we focus on robustness of latent representations as a prerequisite for interpretability.

\begin{ack}
This work was supported by the Cold Spring Harbor Laboratory School of Biological Sciences (J.H.W.), the Jenny and Jeff Kelter Neuroscience Scholarship (J.H.W.), NIH grant RF1DA055666 (T.A.E.), Starr Foundation award I15-0037 (J.H.W. and T.A.E.), and NIH grant RF1NS128901 (J.H.W. and T.A.E).  We thank S. Chauvette and I. Timofeev for sharing of electrophysiological data, which are presented in \cite{soltani_sleepwake_2019}. 
\end{ack}

\bibliographystyle{abbrvnat}
\bibliography{references}
\appendix
\section{Appendix}
\subsection{TN-VAE Loss derivation}
Here, we approximate the relationship between $x_t$ and the latent at the next time point, $z_{t+1}$ with $q_\phi$ (encoder). We seek to minimize the KL divergence between this approximate posterior $q$ and $p_\theta(z_{t+1}, x_{t+1})$. 

\begin{align*}
\textrm{KL}(q_\phi(x_{t+1}|x_t) || p_\theta(z_{t+1}|x_{t+1})) &= \textrm{E}_{z \sim q_\phi(z_{t+1}|x_t)}\log q_\phi(z_{t+1}|x_t) - \textrm{E}_{z \sim q_\phi(z_{t+1}|x_t)}\log p_\theta(z_{t+1}|x_{t+1}) \\
&= \textrm{E}_{z \sim q_\phi(z_{t+1}|x_t)}\log q_\phi(z_{t+1}|x_t) - \textrm{E}_{z \sim q_\phi(z_{t+1}|x_t)}\log p_\theta(x_{t+1}|z_{t+1})
\\ &- \textrm{E}_{z \sim q_\phi(z_{t+1}|x_t)}\log p_\theta(z_{t+1}) + \textrm{E}_{z \sim q_\phi(z_{t+1}|x_t)}\log p_\theta (x_{t+1})
\end{align*}

Minimizing this quantity is equivalent to maximizing the Evidence Lower Bound ($\mathscr{L}$).

\begin{align*}
\mathscr{L} &= \textrm{E}_{z \sim q_\phi(z_{t+1}|x_t)}\log p_\theta(x_{t+1}|z_{t+1}) + \textrm{E}_{z \sim q_\phi(z_{t+1}|x_t)}p_\theta(z_{t+1}) - \textrm{E}_{z \sim z_{t+1}~q_\phi(z_{t+1}|x_t)} \log q_\phi(z_{t+1}| x_t) \\
&= \textrm{E}_{z \sim q_\phi(z_{t+1}|x_t)}\log p_\theta(x_{t+1}|z_{t+1}) - \textrm{KL}(q_\phi(z_{t+1}|x_t)||p_\theta(z_{t+1}))
\end{align*}

\subsection{Neighbor Loss as the likelihood of a random walk in the latent space} 
\label{sec::appa}
\noindent Assume data-points traverse a low-dimensional latent space where $\textbf{x}_{t+1}$ is drawn from a circular Gaussian distribution centered at $\textbf{x}_t$ with variance $\sigma$.
The log-likelihood of this function is:
\begin{align*}
\log(L) &= \log\left(\prod_{t=1}^{n}\frac{1}{2\pi\sigma}\exp(-\frac{(\textbf{x}_{t+1}-\textbf{x}_t)(\textbf{x}_{t+1}-\textbf{x}_t)^T}{2\sigma})\right) \\
&= \sum_{t=1}^{n}\log\left(\frac{1}{2\pi\sigma}\exp(-\frac{(\textbf{x}_{t+1}-\textbf{x}_t)(\textbf{x}_{t+1}-\textbf{x}_t)^T}{2\sigma} )\right) \\
&= -\frac{n}{2}\log 2\pi - n\log\sigma - \frac{1}{2\sigma}\sum[(\textbf{x}_{t+1}-\textbf{x}_t)(\textbf{x}_{t+1}-\textbf{x}_t)^T] .
\end{align*} \\
Thus, maximizing the log-likelihood is equivalent to minimizing sum of the absolute distance between each point and its neighboring point in time.

\setcounter{figure}{0}
\renewcommand{\figurename}{Fig.}
\renewcommand{\thefigure}{A\arabic{figure}}

\begin{figure}[ht]
\begin{center}
\centerline{\includegraphics[width=\columnwidth]{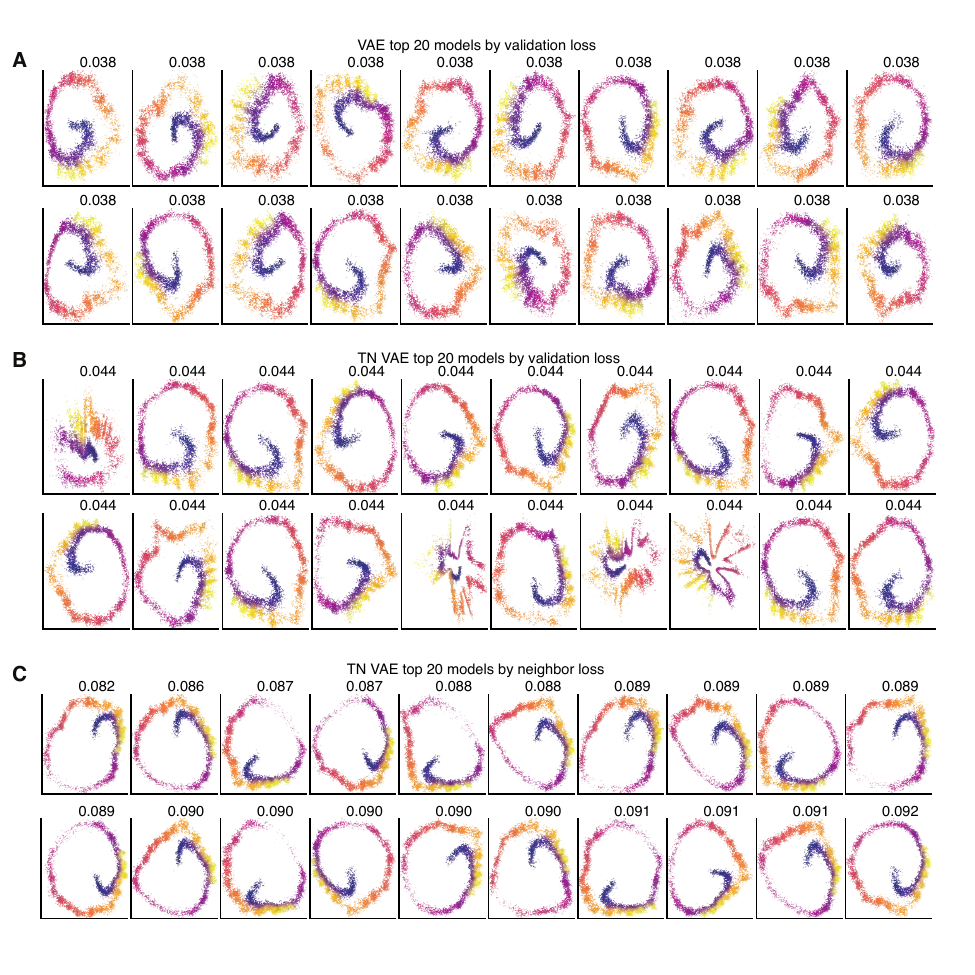}}
\caption{Latent representations of the spiral dataset on a held-out test set from the top 20 models learned by (A) VAE and selected based on the validation loss, (B) TN-VAE and selected based on the validation loss, (C) TN-VAE and selected based on the neighbor loss.}
\label{figs1}
\end{center}
\vskip -0.2in
\end{figure}

\begin{figure}[ht]
\begin{center}
\centerline{\includegraphics[width=\columnwidth]{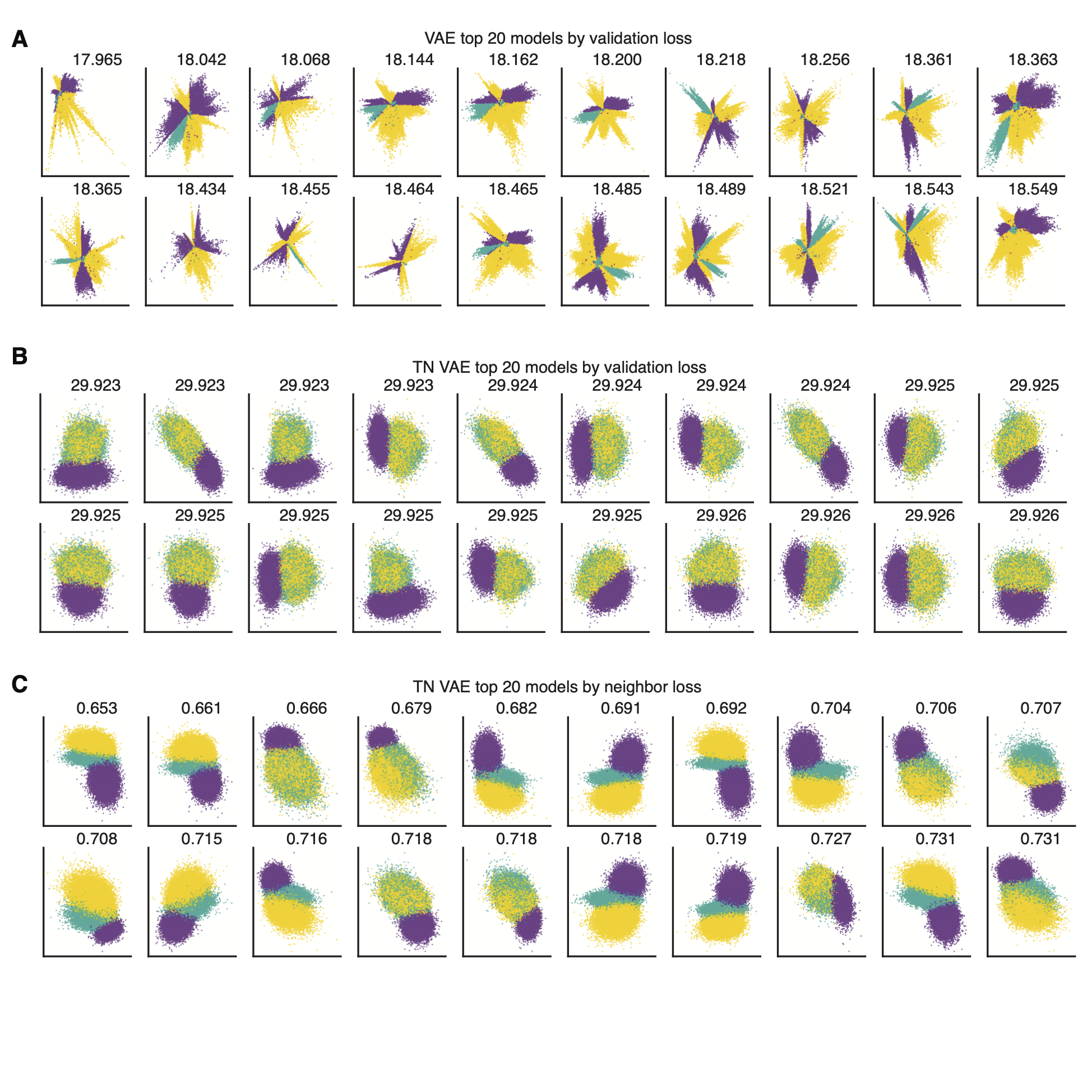}}
\caption{Latent representations of the HMM dataset on a held-out test set from the top 20 models learned by (A) VAE and selected based on the validation loss, (B) TN-VAE and selected based on the validation loss, and (C) TN-VAE and selected based on the neighbor loss.}
\label{figs2}
\end{center}
\vskip -0.2in
\end{figure}

\begin{figure}[ht]
\begin{center}
\centerline{\includegraphics[width=\columnwidth]{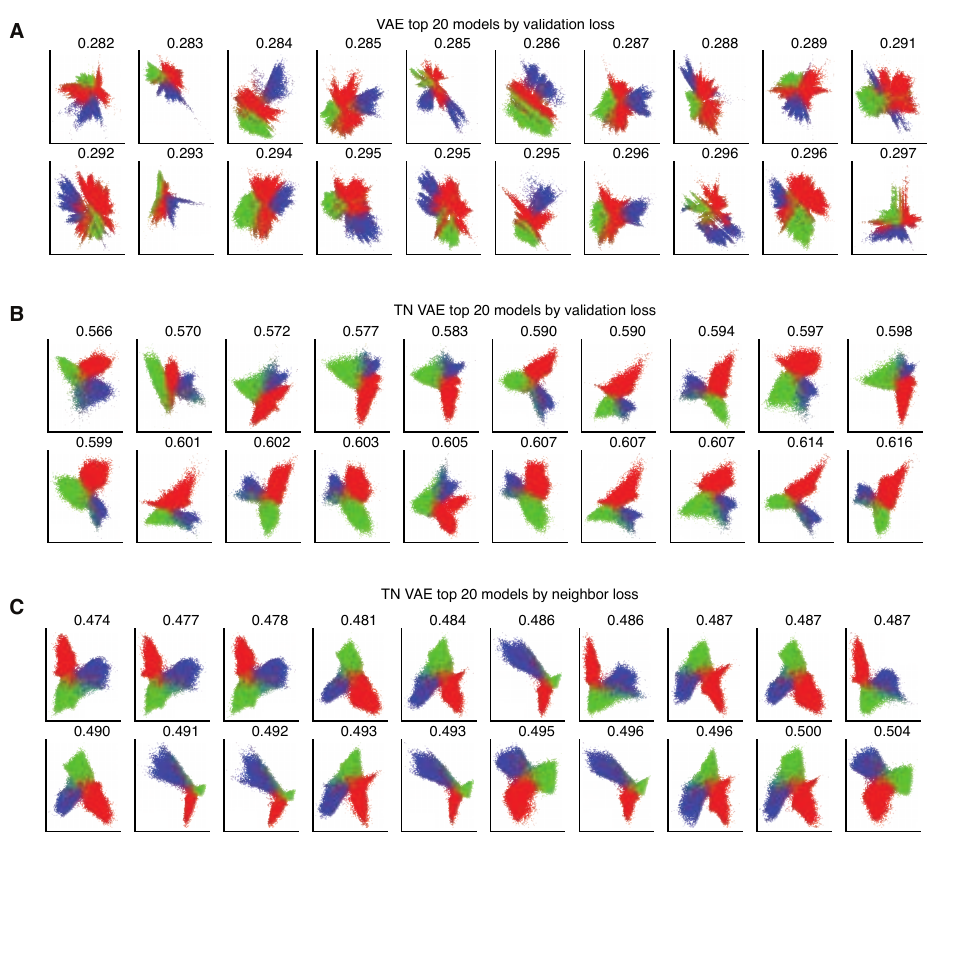}}
\caption{Latent representations of the LFP dataset on a held-out test set from the top 20 models learned by (A) VAE and selected based on the validation loss, (B) TN-VAE and selected based on the validation loss, and (C) TN-VAE and selected based on the neighbor loss.}
\label{figs3}
\end{center}
\vskip -0.2in
\end{figure}

\begin{figure}[ht]
\begin{center}
\centerline{\includegraphics[width=\columnwidth]{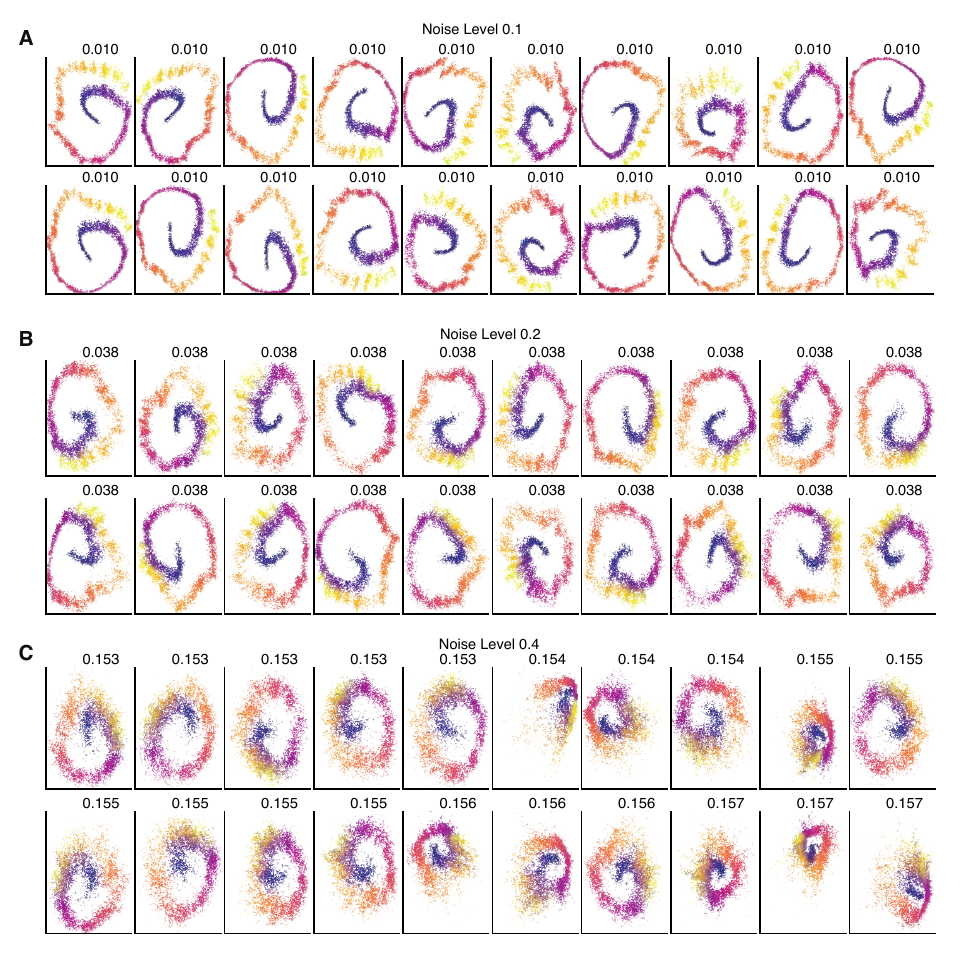}}
\caption{The level and pattern of overfitting in the VAE depends on the magnitude of the high-dimensional noise in the observations. Latent representations of the spiral dataset on a held-out test set from the top 20 models learned by VAE and selected based on the validation loss, for training and test datasets with noise level (A) 0.1, (B) 0.2, and (C) 0.4.}
\label{figs4}
\end{center}
\vskip -0.2in
\end{figure}

\end{document}